\documentclass{article}
\usepackage{wrapfig}
\PassOptionsToPackage{numbers, sort}{natbib}

\usepackage[preprint]{neurips_2026}
\usepackage[utf8]{inputenc} 
\usepackage[T1]{fontenc}    
\usepackage{hyperref}       
\usepackage{url}            
\usepackage{booktabs}       
\usepackage{amsmath}        
\usepackage{amsfonts}       
\usepackage{graphicx}       
\usepackage{nicefrac}       
\usepackage{microtype}      
\usepackage{xcolor}         
\usepackage{booktabs}
\usepackage{multirow}
\usepackage{makecell}

\newcommand{\ie}{\emph{i.e.,}}
\title{SimSD: Simple Speculative Decoding in Diffusion Language Models}

\newcommand{\blfootnote}[1]{%
  \begingroup
  \renewcommand\thefootnote{}\footnotetext{#1}%
  \addtocounter{footnote}{-1}%
  \endgroup
}

\newcommand{\authshift}{\hspace*{-2em}}
\author{
\authshift\textbf{Junxia Cui}\textsuperscript{1}\thanks{Equal contribution.}
\quad \textbf{Haotian Ye}\textsuperscript{1}\footnotemark[1]
\quad \textbf{Runchu Tian}\textsuperscript{2}\footnotemark[1]
\quad \textbf{Hongcan Guo}\textsuperscript{1}
\quad \textbf{Jinya Jiang}\textsuperscript{1}
\quad \textbf{Haoru Li}\textsuperscript{1}\\
\authshift\textbf{Chaojie Ren}\textsuperscript{1}
\quad \textbf{Yiming Huang}\textsuperscript{1}
\quad \textbf{Kaijie Zhu}\textsuperscript{4}
\quad \textbf{Zhongkai Yu}\textsuperscript{1}
\quad \textbf{Kun Zhou}\textsuperscript{1,$\dagger$}
\quad \textbf{Jingbo Shang}\textsuperscript{1,$\dagger$}\\
\authshift\textsuperscript{1}University of California San Diego
\quad
\textsuperscript{2}University of Illinois Urbana-Champaign\\
\authshift\textsuperscript{3}University of California Santa Barbara\\
\authshift\textsuperscript{$\dagger$}Corresponding authors.\\
\authshift\texttt{jucui@ucsd.edu, jshang@ucsd.edu}
}

\begin{document}

\maketitle

\blfootnote{Code is publicly available at \url{https://github.com/airevo2/SimSD-release}.}

\begin{abstract}
Diffusion large language models (dLLMs) have recently emerged as a promising alternative to autoregressive (AR) LLMs, offering faster inference through parallel or blockwise decoding. However, their masked language modeling formulation remains incompatible with standard token-level speculative decoding, one of the most effective acceleration techniques for AR models. In AR decoding, the causal mask preserves temporally valid token-level contexts, enabling a target model to verify multiple drafted tokens in a single forward pass. In contrast, dLLMs rely on mask tokens and bidirectional attention, causing the effective context to change across denoising steps and preventing direct token-level speculative verification. To bridge this gap, we propose a simple but effective speculative decoding algorithm for diffusion language models, named SimSD, which mainly adopts a plug-and-play masking strategy that equips dLLMs with temporally valid token-level contexts for speculative decoding. Our method explicitly introduces reference tokens from draft-model predictions and designs an attention mask that regulates their interaction with current-step tokens, allowing dLLMs to compute valid logits for drafted tokens in a single forward pass. This restores the key verification ability provided by causal masking in AR models while preserving the parallel decoding advantages of dLLMs. The proposed method is training-free and can be flexibly integrated with other acceleration techniques such as KV cache and blockwise decoding. Experiments on SDAR-family dLLMs across four benchmarks show that our method achieves up to \(7.46\times\) higher decoding throughput while maintaining and even improving average generation quality.
\end{abstract}

\section{Introduction}
Large language models (LLMs)~\citep{chowdhery2022palmscalinglanguagemodeling, hurst2024gpt, comanici2025gemini} have achieved remarkable success across a wide range of natural language processing tasks~\citep{zhao2023survey, minaee2024large}. These advances are largely driven by autoregressive (AR) Transformer architectures~\citep{vaswani2017attention}. Despite the performance, AR decoding is inherently sequential, which requires strictly performing token-by-token generation~\citep{fu2024lookahead_decoding, xia-etal-2024-unlocking}.
As an alternative, diffusion large language models (dLLMs)~\citep{NEURIPS2021_958c5305, Diffusion-LM} generate text by iteratively denoising an incomplete sentence with bidirectional attention. With a properly designed denoising schedule, dLLMs enable predicting multiple tokens at one decoding step, leading to promising parallelism and lower inference latency.

Recent commercial dLLMs~\citep{labs2025mercuryultrafastlanguagemodels,gemini_diffusion,bie2025llada2} have substantially narrowed the performance gap with state-of-the-art AR models, while demonstrating much faster inference speed with parallel or blockwise decoding~\citep{christopher2025specdiff, cheng2025sdar, pan2025blockspec, chen2026dflash}. However, further improving dLLM inference speed remains challenging because its masked language modeling design~\citep{devlin2019bert,nie2025llada} is not naturally compatible with many acceleration techniques developed for AR models, especially for the widely-used speculative decoding method~\citep{xia-etal-2024-unlocking,leviathan2023fast,chen2023speculative_sampling}. 
In AR models, the token-level causal mask ensures that every generated token remains in the context for subsequent positions. Thus, for speculative decoding in AR models, all the drafted tokens from a smaller model only require a single forward pass over their concatenated sequence to produce valid logits for all drafted tokens under their correct prefixes. This property is central to speculative decoding, where multiple possible tokens can be verified in parallel. By contrast, dLLMs rely on mask tokens to indicate unresolved positions and use bidirectional attention to condition on the visible context. Consequently, the effective input context changes across denoising or generation steps, and dLLMs cannot maintain temporally valid token-level contexts. Therefore, standard token-level speculative verification cannot be applied directly to dLLMs.

To address this limitation, we propose a simple but effective method namely \textbf{SimSD} that adopts a plug-and-play masking strategy that equips dLLMs with temporally valid token-level contexts for speculative decoding. The key idea is to construct a token-level causal structure in which each token can directly attend to other information from the additional reference context, \ie inference results from a smaller model, analogous to how AR models read in-context tokens through the causal mask. 
To this end, we explicitly add several tokens as the reference context and design an attention mask that regulates interactions between these reference context and the tokens predicted at the current step. In this way, the visible context is no longer determined only by the current masked input, but built to include both current-step tokens and reference context. This restored temporal context allows a dLLM to compute logits for drafted tokens and thereby supports standard speculative verification in a single forward pass. It matches the verification mechanism used in AR speculative decoding, and gives dLLMs the key ability while retaining the parallel decoding benefits.

Importantly, the proposed mask requires no additional training and can be directly applied to existing dLLM decoding pipelines. Since we only modify the design of attention mask and copy few hidden states, our method can be flexibly integrated with other acceleration techniques, such as KVCache and blockwise decoding.
To verify it, we conduct extensive experiments on SDAR~\citep{cheng2025sdar} family dLLMs and four benchmarks, GSM8K~\citep{cobbe2021trainingverifierssolvemath}, MBPP~\citep{austin2021program}, TriviaQA~\citep{joshi-etal-2017-triviaqa}, and MMLU~\citep{hendrycks2020measuring}. Experimental results have demonstrated that our method can achieve up to \(7.46\times\) higher decode throughput, and the generation quality does not degrade and even slightly boosts (+1.7\%), compared with the Vanilla SDAR baseline with tensor parallelism.

\begin{figure}[t]
    \centering
    \resizebox{1.0\linewidth}{!}{
        \includegraphics{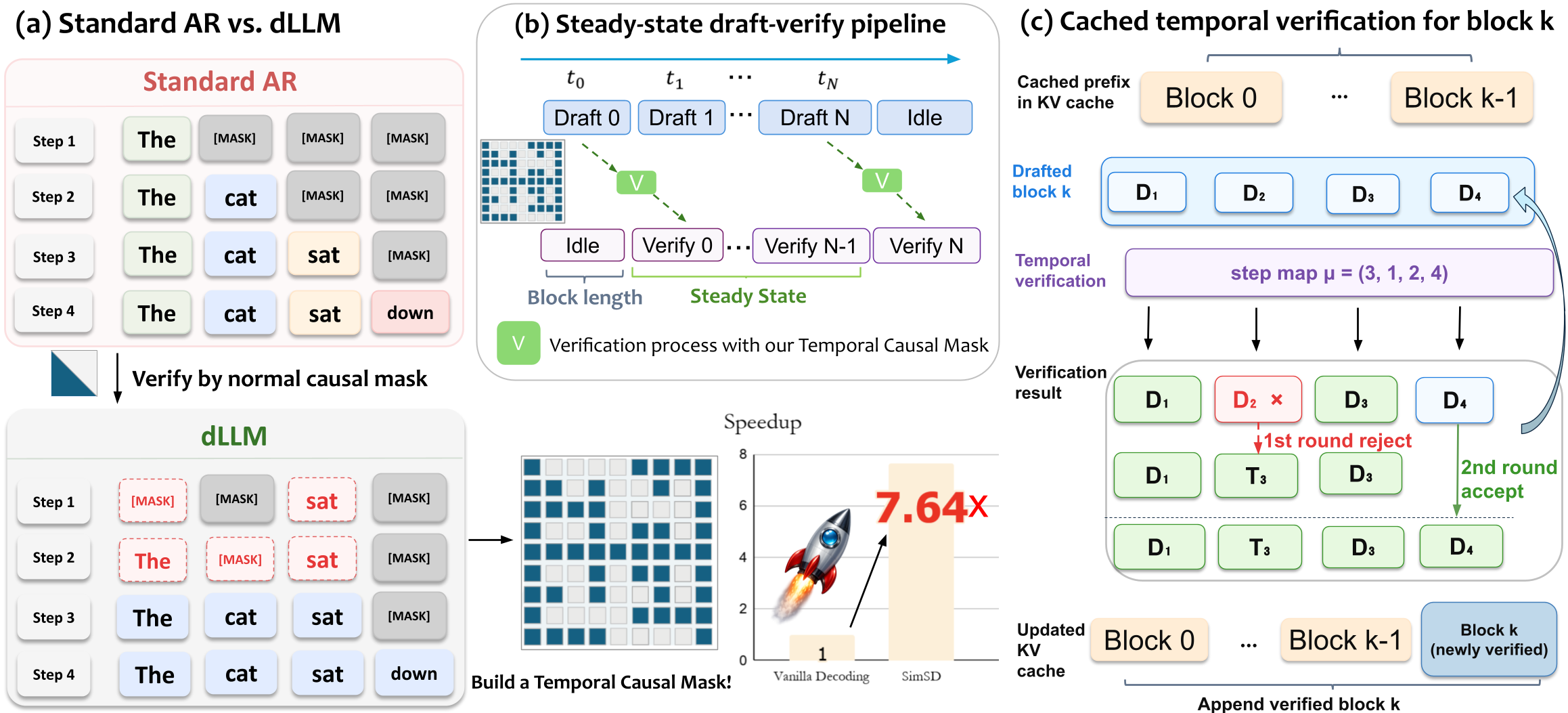}
    }
    \caption{SimSD restores token-level speculative decoding for diffusion language models.
Vanilla dLLMs cannot directly support speculative decoding because bidirectional attention breaks temporal token-level contexts. SimSD adopts a temporal causal mask for supporting standard speculative verification in a single forward pass, leading to faster decoding without hurting performance.
}
    \label{fig:example}
\end{figure}

\section{Related work}

\paragraph{Diffusion Language Models.}
Masked language modeling was first established as a pretraining objective by BERT style bidirectional encoders\citep{devlin2019bert}, which recover a small fraction of held out tokens from their bidirectional context. Subsequent work recast this objective as a continuous time generative process, with DiffusionBERT~\citep{he-etal-2023-diffusionbert} providing an explicit bridge between BERT style masked prediction and absorbing state diffusion. Building on this masked language modeling tradition, discrete diffusion language models~\citep{nie2025llada, yang2025mmada} corrupt a clean sample by replacing a randomly chosen subset of its tokens with the special symbol $\texttt{[MASK]}$, producing a noised sequence. The forward corruption factorizes across all response token positions: each position is independently replaced by $\texttt{[MASK]}$ with a certain probability.
The reverse process supports an accelerated parameterization in which several masked positions are recovered jointly within a single transition from high noise level to a lower level, rather than one position at a time.



\paragraph{Blockwise Decoding for dLLMs.}
A complementary line of work adapts pretrained autoregressive checkpoints into diffusion language models, preserving the AR backbone's linguistic competence while enabling parallel decoding~\citep{ye2025dream7b}. A common strategy is to introduce block causal attention, which imposes causal dependencies across coarse-grained blocks while allowing bidirectional denoising within each block. This design has been adopted by recent dLLMs~\citep{ye2025dream7b, gong2025scaling_diffusion_language_models, cheng2025sdar, xie2025dreamcoder}, suggesting that AR pretraining can transfer effectively to masked diffusion objectives when the attention pattern is properly constrained.
Block diffusion models~\citep{arriola2025block, cheng2025sdar} formalize this idea by partitioning the sequence into contiguous fixed-size blocks and replacing full bidirectional attention with block causal attention. During inference, all positions within a block can be denoised jointly, preserving the parallel sampling advantage of diffusion, while dependencies across blocks remain causal. This structure also enables standard KV caching across blocks, avoiding the cache invalidation issues caused by combining bidirectional attention with KV reuse in full attention dLLMs~\citep{hu2025acceleratingdiffusionlanguagemodel, liu2025dllmcache, ma2025dkvcachecachediffusionlanguage, arriola2025block, wu2025fastdllm}.

\paragraph{Speculative Decoding.}
Speculative decoding (SD) accelerates LLM inference by using a small draft model to propose multiple candidate tokens and a larger target model to verify them in parallel~\citep{leviathan2023fast, chen2023speculative_sampling}. This paradigm is particularly effective at low batch sizes on GPUs, where decoding is often memory bandwidth bound rather than compute bound. 
Subsequent work improves this pipeline through better proposal models, tree-based verification, multi-token prediction heads, feature level drafting, and drafter-free parallel decoding~\citep{miao2023specinfer, cai2024medusa, li2024eagle, fu2024lookahead_decoding}. 
Existing work on diffusion-based speculative decoding falls into two main directions. The first uses diffusion models as drafters for autoregressive verifiers. SpecDiff and SpecDiff-2 train or align diffusion drafters to improve proposal quality for AR targets~\citep{christopher2025specdiff, sandler2025specdiff2}; DiffuSpec instead reuses a pretrained diffusion language model as a training-free drafter while retaining an AR verifier~\citep{li2025diffuspec}, and DFlash explores block diffusion drafting under the same AR target setting~\citep{chen2026dflash}. 
The second direction accelerates a single dLLM through self-speculation or trajectory-level verification, where the same diffusion model or block diffusion process proposes and checks candidate states~\citep{gao2025ssd, han2026s2d2, agrawal2025spiffy, pan2025blockspec}. For example, S2D2 is described as a training-free self-speculative decoding framework for block-diffusion language models, where the same pretrained model serves as both drafter and verifier under different modes; BlockSpec similarly targets blockwise speculative decoding for dLLMs by exploring future decoding trajectories. 

\section{Methodology}

As dLLMs can not maintain temporally valid token-level contexts as AR models to support multiple possible tokens verification in a single forward pass, we aim to devise a new attention masking strategy to solve it in dLLMs.
Based on it, we propose a simple but effective speculative decoding method for accelerating dLLMs during inference, which can significant reduce the inference latency and support the integration with other techniques such as blockwise decoding and KV cache.

\subsection{Token-level Temporal Causal Attention}
\label{sec:Draft model}
Original dLLMs use bidirectional attention to control the visible context. Here, we replace this design with a token-level temporal causal attention strategy, which ensures that each token can attend to its current step context as well as additional reference information from earlier temporal steps. Concretely, we first construct an input layout that keeps copies of related tokens as reference context, then build an attention mask according to token-level temporal order, and finally align position encodings before predicting the masked tokens.

\paragraph{Input Layout Prepare.}
Given a prompt $c$, we construct the attention context by appending two response sections of length $B$: a data section $D=(D_1,\dots,D_B)$ and a mask section $\widetilde{D}=(\widetilde{D}_1,\dots,\widetilde{D}_B)$. The data section stores reference tokens, while the mask section contains the prediction slots, where $\widetilde{D}_k=\texttt{[MASK]}$. Each pair $(D_k,\widetilde{D}_k)$ corresponds to the same response position but plays different roles: $D_k$ provides contextual information from a reference or previously decoded trajectory, whereas $\widetilde{D}_k$ is the position where logits are computed for token prediction. This layout allows the model to explicitly keep reference tokens in the context while still performing masked token prediction.

\paragraph{Temporal Attention Mask.}
Each response position is assigned a temporal step label \(\tau \in \{1,\dots,B\}\), indicating the inference step associated with that token. We represent this assignment using a step map \(\mu=(\mu_1,\dots,\mu_B)\), a permutation of \(\{1,\dots,B\}\), where both \(D_k\) and \(\widetilde{D}_k\) share the same temporal label \(\tau=\mu_k\). Prompt tokens are assigned \(\tau=0\). For example, the left-to-right order corresponds to \(\mu=(1,2,\dots,B)\), while other permutations allow different decoding orders. This step map extends block-level causal attention~\citep{arriola2025block} to a finer token-level temporal structure within each block.
We define the attention mask \(A\) by \(A_{ij}=1\) if position \(i\) is allowed to attend to position \(j\), and \(A_{ij}=0\) otherwise. Prompt tokens are visible to all response positions. For response positions, the mask is defined as
\begin{equation}
\small
A_{ij} =
\begin{cases}
1 & \text{if } i,j \in D \text{ and } \tau(j) \le \tau(i), \\
1 & \text{if } i \in D,\, j \in \widetilde{D} \text{ and } \tau(j) > \tau(i), \\
1 & \text{if } i \in \widetilde{D},\, j \in D \text{ and } \tau(j) < \tau(i), \\
1 & \text{if } i,j \in \widetilde{D} \text{ and } \tau(j) \ge \tau(i), \\
0 & \text{otherwise.}
\end{cases}
\end{equation}
This mask enforces a temporally valid visible context. A masked prediction slot \(\widetilde{D}_i\) can attend to decoded or reference data tokens from earlier steps, \(D_j\) with \(\tau(j)<\tau(i)\), while future or current unresolved positions remain represented by mask placeholders. Conversely, a data token \(D_i\) is computed from the information that would have been available when it became known: earlier data tokens and later mask placeholders. Padding positions are masked out by zeroing their corresponding rows and columns.

\paragraph{Position Encoding Alignment.}
Because the data and mask sections contain paired positions, their positional encodings must be aligned. Under the above layout, \(D_k\) and \(\widetilde{D}_k\) refer to the same response position but appear at different absolute sequence indices. With standard rotary position embedding (RoPE)~\citep{su2021roformer}, this would rotate their query and key vectors using different position indices, even though they correspond to the same token position. To avoid this mismatch, we apply a RoPE copy strategy: each mask position \(\widetilde{D}_k\) reuses the RoPE index of its paired data position \(D_k\), while prompt and data tokens keep their original absolute indices. This alignment ensures that logits computed at masked positions are positionally consistent with the corresponding response tokens.

\paragraph{Masked Token Prediction.}
After applying the temporal attention mask and position encoding alignment, the model predicts tokens only at the mask section. For a prediction slot \(\widetilde{D}_i\), the visible context consists of the prompt, earlier data tokens \(D_j\) with \(\tau(j)<\tau(i)\), and mask placeholders \(\widetilde{D}_j\) with \(\tau(j)\ge\tau(i)\). Therefore, the logits at \(\widetilde{D}_i\) approximate the token distribution under a temporally consistent context, analogous to the prefix-conditioned distribution used in autoregressive speculative verification. This enables dLLMs to verify drafted tokens at the token level while preserving their parallel decoding structure.

\begin{figure}[t]
  \centering
  \includegraphics[width=1\linewidth]{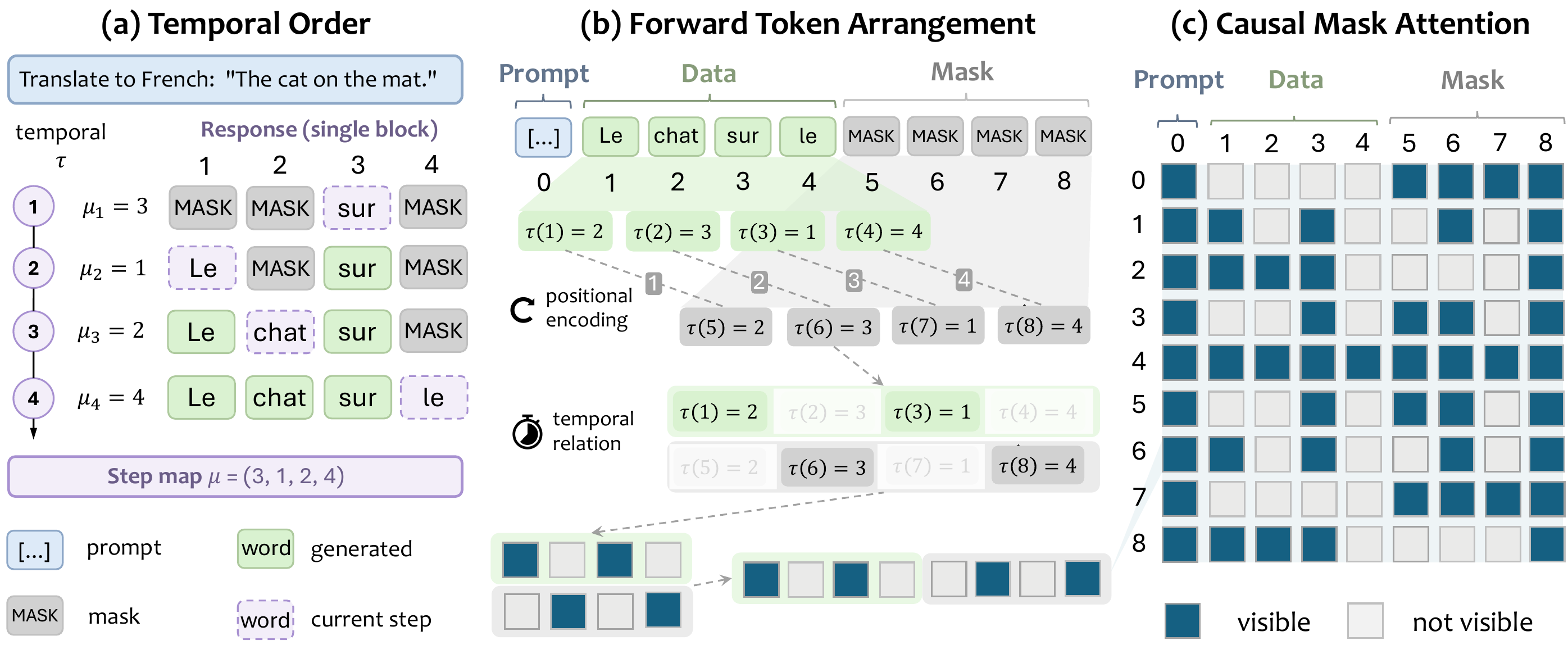}
  \caption{Token-level temporal causal attention. (a) A step map assigns each token a temporal decoding order. (b) The input is arranged as prompt, data, and mask tokens, with shared temporal labels and aligned positional encoding. (c) The attention mask enforces token-level temporal causality.}
  \label{fig:standard-vs-causal}
\end{figure}

\subsection{Application for Speculative Decoding}
\label{sec:speculative-decoding}

We next describe how token-level temporal causal attention enables standard speculative decoding for dLLMs. Speculative decoding typically uses a small draft model to propose candidate tokens and a larger target model to verify them in parallel with rejection sampling~\citep{leviathan2023fast, chen2023speculative_sampling}. In our setting, both the drafter and verifier are dLLMs: the smaller dLLM first proposes candidate tokens together with their temporal order, and the target dLLM then scores all candidates in a single forward pass using the proposed token-level temporal causal mask.


\paragraph{Draft Model for Candidates Collection.}

Consider a dLLM decoding sequence of length $B$, and let $R$ denote the number of remaining undecoded positions. When $R \ge \gamma$, a smaller draft dLLM with parameters $\phi$ proposes $\gamma$ candidate tokens $\widehat{D} = (\widehat{D}_{o_1}, \ldots, \widehat{D}_{o_\gamma})$, where $o_1,\ldots,o_\gamma$ denotes the draft-induced temporal order. Specifically, position $o_r$ is generated at draft step $r$. The drafter therefore determines both the candidate tokens and the order in which they should be verified. The target verifier reuses this order by assigning temporal label $\tau=r$ to the slot corresponding to $\widehat{D}_{o_r}$.

\paragraph{Target Model Verification.}
The verification input follows the paired data-mask layout introduced in Sec.~\ref{sec:Draft model}, but only for the $\gamma$ drafted positions in the current speculative step. The $r$-th data slot stores the drafted token $\widehat{D}_{o_r}$, and the paired mask slot serves as its verification position. We then construct the temporal attention mask using $\tau=r$ for each drafted token and apply the same RoPE-copy alignment between each drafted token and its paired verification slot.
Under this mask, the verification slot for $\widehat{D}_{o_r}$ can attend to the prompt and to earlier drafted tokens $\widehat{D}_{o_1},\ldots,\widehat{D}_{o_{r-1}}$, but not to the current token or later drafted tokens. Thus, the target distribution at this slot, denoted by $p_\theta^{(r)}$, scores $\widehat{D}_{o_r}$ under the valid draft-induced causal context. Importantly, all target distributions $\{p_\theta^{(r)}\}_{r=1}^{\gamma}$ are obtained from one target forward pass, allowing the target dLLM to verify multiple drafted tokens in parallel while preserving token-level causal validity.
We apply the standard speculative decoding acceptance rule along the draft-induced temporal order. Let $p_\phi^{(r)}$ be the draft distribution used to sample $\widehat{D}_{o_r}$. For $r=1,\ldots,\gamma$, the candidate token is accepted with probability
\begin{equation}
\small
\alpha_r
=
\min\left(
1,
\frac{
p_{\theta}^{(r)}(\widehat{D}_{o_r})
}{
p_{\phi}^{(r)}(\widehat{D}_{o_r})
}
\right).
\end{equation}
The algorithm accepts the longest prefix before the first rejection. If $\widehat{D}_{o_r}$ is rejected, the remaining drafted suffix is discarded, and the target dLLM  samples the token at position $o_r$. The next speculative step then continues from the updated block state. If all $\gamma$ candidates are accepted, they are committed to the block, and speculative decoding proceeds while at least $\gamma$ undecoded positions remain. When fewer than $\gamma$ positions remain, drafting stops and the target dLLM completes the block directly.

\paragraph{Training-free Deployment.}
The proposed procedure is training-free. It only modifies the verification-time input layout, attention mask, and RoPE indexing, and can therefore be applied to pretrained draft and target dLLMs without parameter updates. The rejection-sampling correction preserves the target distribution induced by the token-level temporal causal mask, while reducing the number of expensive target forward passes. This makes cross model dLLM-to-dLLM speculative decoding practical without jointly designing or retraining the drafter and verifier.

\subsection{Efficiency Discussion} 

Consider one speculative verification step with draft length $\gamma$. Let $T_\theta$ be the walltime of one target dLLM verification pass, and let $T_\phi$ be the average walltime of one draft step. One speculative verification step with draft length $\gamma$ costs, where $\rho$ define the draft-to-target cost ratio:
\begin{equation}
\small
T_{\mathrm{SD}}(\gamma)=\gamma T_\phi + T_\theta=(1+\gamma\rho)T_\theta;~\rho = \frac{T_\phi}{T_\theta}.
\end{equation}
Following the analysis of \citet{leviathan2023fast}, we aim to quantify the benefit of verifying multiple draft tokens with one target forward pass. Our verification layout introduces an additional overhead because it duplicates the drafted sequence to form the data-mask paired input. Let $\ell$ denote the draft sequence length and let $n$ be the original verification sequence length before duplication. The paired layout increases the effective sequence length from $n$ to $n+\ell$. Therefore, the attention cost increases from $O(n^2)$ to $O((n+\ell)^2)$, with an extra cost. 
\begin{equation}
O((n+\ell)^2 - n^2)
=
O(2n\ell + \ell^2).    
\end{equation}

\paragraph{Adaptation under Blockwise Decoding and KV Cache.}
The duplicated verification layout has a bounded overhead under blockwise decoding with KV cache. 
Let \(n\) be the prefix length and \(\ell\) be the duplicated draft length. 
With KV cache, the prefix keys and values are reused, so the verifier only computes the newly introduced duplicated positions. 
Thus the additional computation is bounded by
\begin{equation}
O((n+\ell)^2-n^2)=O(n\ell+\ell^2) \le O(nB+B^2).
\end{equation}
where the inequality follows from \(\ell \le B\) under blockwise decoding. 
Since \(B\) is a small constant in our experiments, e.g., \(B=4\) or \(B=8\), the overhead is effectively linear in the cached prefix length and is much smaller than duplicating and recomputing the full response sequence. In practical settings, this overhead is acceptable.
\section{Experiments}

\subsection{Experimental setup}

\paragraph{Models.} Following previous studies\citep{leviathan2023fast}\citep{chen2023speculative_sampling}, we use small scale draft model and large scale target model from the same model series for the speculative decoding experiments. We use dLLMs trained with different block lengths from the widely used SDAR\citep{cheng2025sdar} series.

\paragraph{Datasets \& metrics.}
To assess the generation efficiency and effectiveness of our method, we evaluate it on four standard benchmarks covering different task categories: \textbf{GSM8K}~\citep{cobbe2021trainingverifierssolvemath} for mathematical reasoning, \textbf{MBPP}~\citep{austin2021program} for code generation, and \textbf{TriviaQA}~\citep{joshi-etal-2017-triviaqa} and \textbf{MMLU}~\citep{hendrycks2020measuring} for general NLP and knowledge-intensive evaluation. For efficiency, we report \textbf{TPS} and the \textbf{speedup ratio} against vanilla decoding. For effectiveness, we report accuracy for GSM8K, TriviaQA, and MMLU, and pass@1 for MBPP.

\paragraph{Baselines.}
(i) \textbf{Vanilla} follows the standard dLLM decoding procedure with block causal attention and KV cache. Since speculative decoding uses two GPUs, we run Vanilla with tensor parallelism = 2 for a fair comparison. 
(ii) \textbf{Vanilla with CUDA Graph} augments Vanilla decoding with CUDA Graphs to reduce kernel launch overhead. 
(iii) \textbf{S2D2}~\citep{han2026s2d2} performs training free self speculation, where the dLLM first drafts tokens and then verifies them from left to right using the same model with block size set to 1. SSD~\citep{gao2025ssd} is designed for fully bidirectional dLLMs such as LLaDA and Dream, while our experiments use SDAR~\citep{cheng2025sdar}, a block causal dLLM models series with KV cache. BlockSpec~\citep{pan2025blockspec} is orthogonal to our verifier-side contribution, as it focuses on blockwise trajectory speculation rather than token-level target verification. Therefore, these methods are not direct baselines under our SDAR-based setting, and we do not include them in the main comparison.

\paragraph{Configurations.}

We follow the default prompts and chat templates from the OpenCompass~\citep{2023opencompass} evaluation framework. 
Unless otherwise specified, we fix the maximum generation length to \(L=512\) and set the number of denoising steps to the response sequence length \(N\). 
For block length \(B=4\), we use SDAR-1.7B-Chat as the draft model and SDAR-8B-Chat as the target model; for block length \(B=8\), we use SDAR-1.7B-Chat-b8 and SDAR-8B-Chat-b8. 
We set the speculative draft length to \(\gamma=B\), and fall back to vanilla target decoding when the number of remaining positions in the current block satisfies \(R<\gamma\). 
All generation-quality evaluations are conducted in the zero-shot setting with temperature \(=1.0\) and top\_\(p=1.0\). 
All experiments use batch size \(1\), FP32 precision, and KV cache. 
We use two NVIDIA RTX PRO 6000 Blackwell GPUs, with the draft and target models deployed on separate GPUs to avoid resource contention; for the Vanilla baseline, we use tensor parallelism of size \(2\). 
CUDA Graph is enabled for all methods except the basic Vanilla baseline. 
Throughput is measured as decoded tokens per second, excluding prompt prefilling, with \(3\) warmup runs. 
For speed evaluation, we use \(200\) examples from each dataset, while generation quality is evaluated on the full benchmark. 
Speedup is computed relative to Vanilla TP-2 under the same block length.

\subsection{Main results}
\begin{table}[h]
\centering
\small
\caption{Efficiency Performance.
We report decode tokens per second. 
S2D2 denotes self-speculative decoding for diffusion LLMs, and SD denotes our speculative decoding method.
Speedup is measured relative to Vanilla with tensor parallelism of size 2 under the same block size. BL denotes the block length. CG denotes CUDA Graph. Van. denotes vanilla decoding}
\label{tab:sd_pipe_fast_denoise_tp2_speedup}
\setlength{\tabcolsep}{4.5pt}
\renewcommand{\arraystretch}{1.10}

\begin{tabular}{clcccccc}
\toprule
\multirow{2}{*}{\textbf{BL}}
& \multirow{2}{*}{\textbf{Methods}}
& \multicolumn{5}{c}{\textbf{Decoded Tokens per Second}}
& \multirow{2}{*}{\textbf{Speedup}} \\
\cmidrule(lr){3-7}
& & \textbf{GSM8K} & \textbf{TriviaQA} & \textbf{MBPP} & \textbf{MMLU} & \textbf{Mean} & \\
\midrule

\multirow{4}{*}{4}
& Vanilla
& 10.0 & 11.4 & 11.4 & 5.59 & 9.6 & 1.00$\times$ \\
& Vanilla + CG
& 50.1 & 50.5 & 50.7 & 56.04 & 51.8 & 5.40$\times$ \\
& S2D2
& 54.8 & 28.3 & 47.7 & 32.3 & 40.8 & 4.25$\times$ \\
& \textbf{SimSD}
& \textbf{72.6} & \textbf{73.3} & \textbf{72.2} & \textbf{68.4} & \textbf{71.6} & \textbf{7.46$\times$} \\
\midrule

\multirow{4}{*}{8}
& Vanilla
& 15.3 & 20.4 & 18.7 & 6.21 & 15.2 & 1.00$\times$ \\
& Vanilla + CG
& 30.9 & 31.9 & 30.4 & 36.8 & 32.5 & 2.14$\times$ \\
& S2D2
& 57.6 & 30.0 & 50.2 & 28.0 & 41.5 & 2.74$\times$ \\
& \textbf{SimSD}
& \textbf{74.1} & \textbf{88.3} & \textbf{80.4} & \textbf{84.4} & \textbf{81.8} & \textbf{5.40$\times$} \\
\bottomrule
\end{tabular}
\end{table}
\paragraph{Efficiency Performance.}
Table~\ref{tab:sd_pipe_fast_denoise_tp2_speedup} demonstrates the main efficiency comparison between our method and various baselines. Overall, our method consistently outperforms all the other baselines on decoding throughput across all benchmarks and block sizes. Our method reaches \(71.6\) and \(81.8\) tokens/s on average for \(B=4\) and \(B=8\), corresponding to \(7.46\times\) and \(5.40\times\) speedups over Vanilla decoding. Compared with S2D2, our method also provides substantially higher throughput, showing that our construction of temporal causal mask provides more precise context information.
\begin{table}[h]
\centering
\small
\caption{Accuracy performance.
We report accuracy for GSM8K, TriviaQA, and MMLU, and pass@1 for MBPP.
Van. denotes vanilla decoding, SD denotes our method SimSD, BL denotes the block length and $\alpha$ denotes the token-level acceptance rate.}
\label{tab:pass1_sd_vanilla}
\setlength{\tabcolsep}{3pt}
\renewcommand{\arraystretch}{1.10}
\begin{tabular}{ccccccccccccccc}
\toprule
\multirow{2}{*}{\textbf{BL}}
& \multicolumn{3}{c}{\textbf{GSM8K}}
& \multicolumn{3}{c}{\textbf{MBPP}}
& \multicolumn{3}{c}{\textbf{TriviaQA}}
& \multicolumn{3}{c}{\textbf{MMLU}}
& \multicolumn{2}{c}{\textbf{Avg.}} \\
\cmidrule(lr){2-4}
\cmidrule(lr){5-7}
\cmidrule(lr){8-10}
\cmidrule(lr){11-13}
\cmidrule(lr){14-15}
& Van. & SD & $\alpha$
& Van. & SD & $\alpha$
& Van. & SD & $\alpha$
& Van. & SD & $\alpha$
& Van. & SD \\
\midrule
4
& \textbf{0.922} & 0.896 & 0.915
& \textbf{0.669} & 0.662 & 0.806
& 0.486 & \textbf{0.537} & 0.576
& 0.706 & \textbf{0.758} & 0.973
& 0.696 & \textbf{0.713} \\
8
& \textbf{0.906} & 0.894 & 0.867
& \textbf{0.642} & 0.634 & 0.731
& 0.471 & \textbf{0.503} & 0.519
& 0.704 & \textbf{0.760} & 0.949
& 0.681 & \textbf{0.698} \\
\bottomrule
\end{tabular}
\end{table}
\paragraph{Accuracy performance.}
As the results shown in table~\ref{tab:pass1_sd_vanilla}, it is convinced that our speculative decoding method preserves generation quality comparable to, and in some cases slightly better than, vanilla decoding. Across block lengths 4 and 8, the average performance remains on par with vanilla decoding and even shows small improvements. For block length 4, the average score increases from $0.696$ under vanilla decoding to $0.713$ with speculative decoding. Similarly, for block length 8, the average score improves from $0.681$ to $0.698$. Although our attention mask construction introduces a certain approximation to the exact step by step denoising process, these results indicate that the approximation does not noticeably harm task accuracy in the speculative decoding setting. Instead, the proposed verification mechanism improves decoding efficiency while maintaining, and sometimes slightly improving, generation quality.
\subsection{Additional Results}
\paragraph{Precision Analysis of SimSD.}
Since all positions are evaluated in one forward pass, this construction is not an exact lossless replay of sequential denoising. Instead, it provides an approximation of the same temporal conditioning structure. The approximation arises only from higher order interactions among hidden states that are computed simultaneously, while the direct attention visibility of each position matches the intended denoising step.

We use two experiments to quantify the practical effect of this approximation. (i) \textbf{Self draft distribution analysis.} In this setting, we compare the output distributions induced by conventional generation and by SD generation conditioned on the correct tokens, which directly measures the distributional discrepancy introduced by the parallel construction. (ii) \textbf{Cross model distribution analysis.} In the second experiment, we evaluate whether this approximation affects the normal speculative decoding pipeline, where draft tokens are proposed by a smaller model and verified by a larger target model.

\begin{figure}[h]
\centering

\begin{minipage}{0.48\linewidth}
  \centering
  \includegraphics[width=\linewidth]{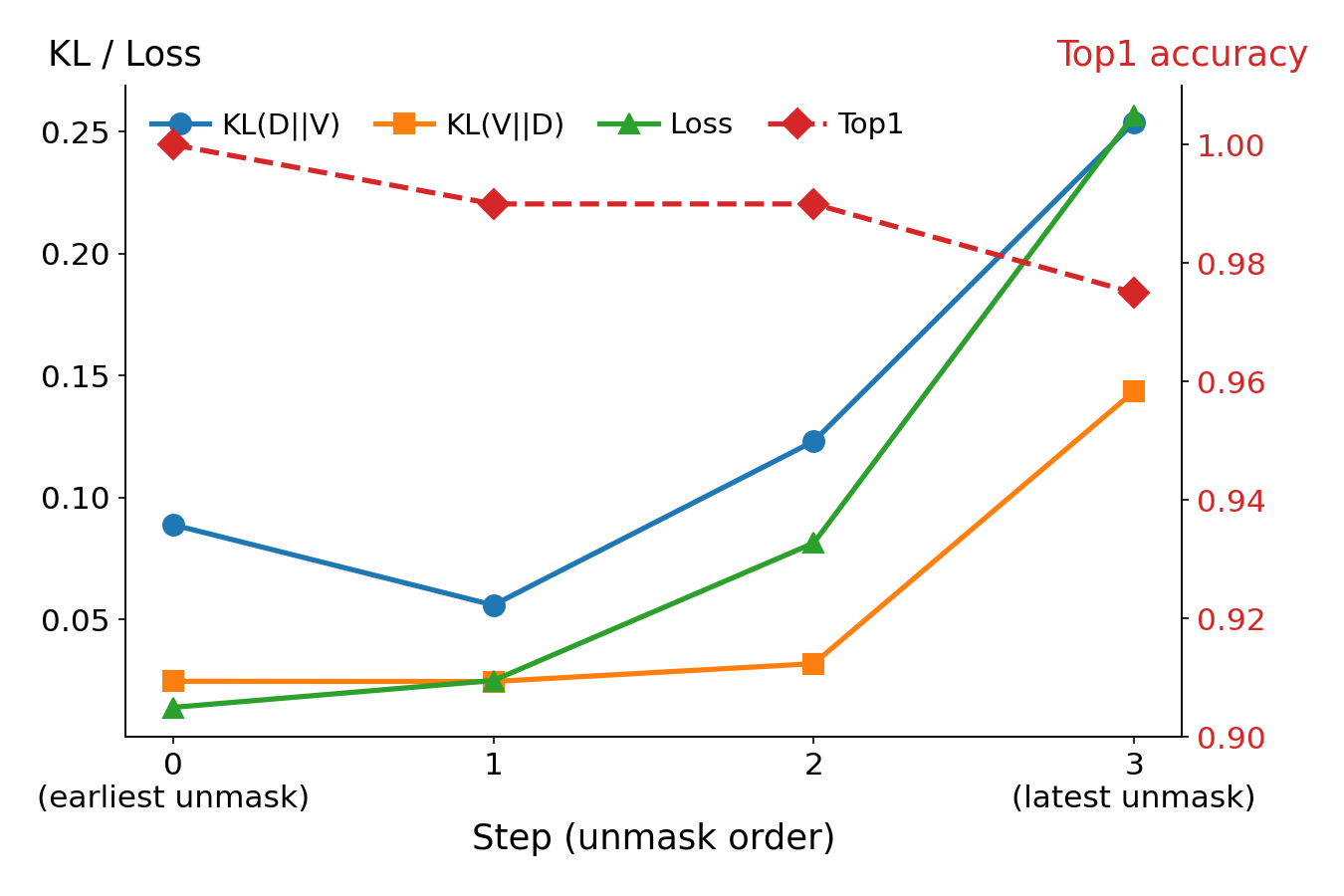}
  \caption{Self draft distribution analysis.
  We use SDAR-8B-Chat as both draft and target with $B=4$ and temperature $0$.}
  \label{fig:self_draft}
\end{minipage}
\hfill
\begin{minipage}{0.48\linewidth}
  \centering
  \includegraphics[width=\linewidth]{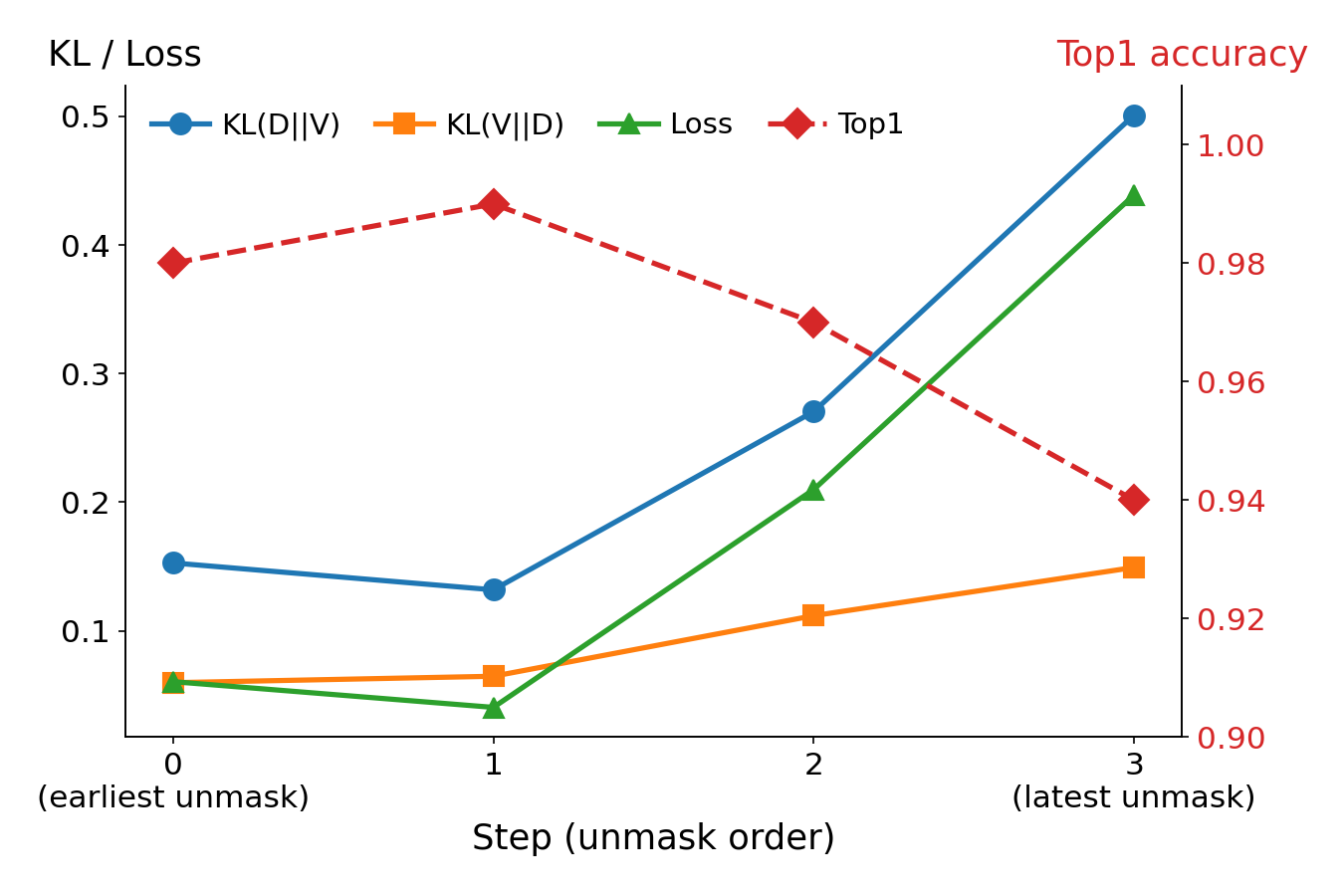}
  \caption{Cross model distribution analysis.
  We use SDAR-1.7B-Chat as draft and SDAR-8B-Chat as target with $B=4$ and temperature $0$.}
  \label{fig:cross_model}
\end{minipage}

\end{figure}

We further analyze the distributional approximation error introduced by the causal mask construction. For both figures, $D$ and $V$ in $\mathrm{KL}(D\|V)$ denote the draft and verification distributions, respectively, and Loss denotes the CE loss. As shown in Figure~\ref{fig:cross_model} and Figure~\ref{fig:self_draft}, the discrepancy between sequential denoising and parallel verification gradually increases from step 0 to step 3. In the self draft setting, $\mathrm{KL}(D\Vert V)$ increases from $0.0889$ at step 0 to $0.2537$ at step 3, while the loss increases from $0.0141$ to $0.2567$. A similar trend appears in the cross model setting, where $\mathrm{KL}(D\Vert V)$ increases from $0.1528$ at step 0 to $0.5010$ at step 3, and the loss increases from $0.0605$ to $0.4389$.

The self draft setting consistently exhibits smaller approximation error than the cross model setting. For example, at step 3, $\mathrm{KL}(D\Vert V)$ is $0.2537$ for self draft but $0.5010$ for cross model, and the corresponding loss is $0.2567$ versus $0.4389$. This is expected because self draft uses the same model on both sides, whereas cross model verification additionally includes the distributional mismatch between the draft and target models. Nevertheless, the overall approximation error remains small in both settings, as reflected by the high top 1 agreement. The self draft setting maintains top 1 agreement of at least $0.9750$, and the cross model setting remains at least $0.9400$ even at the last step. These results indicate that the causal mask provides highly accurate reference context for each verification position, quantitatively explaining why the main accuracy results do not show degradation under our speculative decoding method.
\paragraph{Effect of $\gamma$.}
We analyze how the choice of $\gamma$ affects the quality and efficiency of SimSD. As shown in Table~\ref{tab:gamma_effect_gsm8k}, both the realized acceptance rate and final accuracy generally increase as $\gamma$ becomes larger. This trend suggests that a longer draft scope provides richer target side verification context, allowing SimSD to accept higher quality draft tokens and produce more reliable generations. In particular, increasing $\gamma$ from 3 to 8 improves real $\alpha$ from 0.834 to 0.964 and accuracy from 50.0\% to 82.0\%, showing that larger $\gamma$ values can substantially improve the quality of the draft and verify process.

In contrast, TPS does not exhibit a simple monotonic pattern with respect to $\gamma$. This is expected because changing $\gamma$ affects more than the verification computation itself. A larger $\gamma$ changes the number of tokens verified per step, the interaction between $\gamma$ and the block length, and the overhead of KV cache management during multi block verification. These factors jointly determine the end to end runtime, so the throughput cannot be explained solely by the acceptance rate or the verification scope. For example, $\gamma=4$ achieves the highest TPS in our experiment, while $\gamma=8$ achieves the best accuracy and acceptance rate. This indicates a practical tradeoff: larger $\gamma$ improves generation quality and acceptance behavior, whereas intermediate $\gamma$ values may provide a better speed quality balance. Therefore, we choose $\gamma$ equal to the block length for main experiment, which provides the best overall balance between generation quality and TPS in our setting.

\begin{wraptable}{r}{0.52\textwidth}
\centering
\small
\caption{Effect of $\gamma$ on GSM8K.
We evaluate different values of $\gamma$ using SDAR-1.7B-b8 as draft model and  SDAR-8B-b8 as target model with block length $B=8$ on GSM8K. TPS denotes generated tokens per second,
real-$\alpha$ denotes the realized token-level acceptance rate, and avg accept denotes
the average number of accepted tokens per verification step.}
\label{tab:gamma_effect_gsm8k}
\setlength{\tabcolsep}{4pt}
\renewcommand{\arraystretch}{1.08}
\resizebox{0.52\textwidth}{!}{
\begin{tabular}{cccccc}
\toprule
$\gamma$ & Accuracy & TPS & real-$\alpha$ & Avg. accept & ms/block \\
\midrule
3
& 50.0\% & 55.14 & 0.834 & 4.54/8 & 90 \\
4
& 54.0\% & \textbf{74.40} & 0.887 & 4.93/8 & \textbf{71} \\
5
& 68.0\% & 51.17 & 0.909 & 4.98/8 & 104 \\
6
& 70.0\% & 62.94 & 0.933 & 5.31/8 & 89 \\
7
& 76.0\% & 51.92 & 0.948 & 5.61/8 & 113 \\
8
& \textbf{82.0\%} & 58.74 & \textbf{0.964} & \textbf{6.18/8} & 109 \\
\bottomrule
\end{tabular}
}
\end{wraptable}

\paragraph{Ablation Study on RoPE Alignment.}
To examine the impact of RoPE alignment, we fix the attention structure and vary only the position ids assigned to tokens in the mask block. In the default setting, referred to as Aligned RoPE, each mask token copies the position id of its corresponding token in the reference context, so the two tokens share the same rotary position embedding. In the Without Aligned RoPE setting, mask tokens are instead assigned position ids according to their physical locations in the concatenated sequence, continuing after the reference context rather than reusing the data-token positions.

Table~\ref{tab:share-mask-position-ablation} shows that sharing RoPE positions between each mask token and its paired data token is crucial for valid target verification. Without aligned RoPE, GSM8K accuracy drops from $84.0\%$ to $0.0\%$, while the token-level acceptance rate also decreases substantially from $0.902$ to $0.718$. The large KL divergence of $7.12$ further indicates that physically continued position ids break the intended temporal verification geometry.
\begin{table}[h]
\centering
\small
\caption{RoPE Alignment Ablation.
We evaluate SDAR-1.7B-Chat as the draft model and SDAR-8B-Chat as the target model on GSM8K.
Acc. denotes accuracy, and KL denotes KL divergence averaged over the target output distributions.}
\label{tab:share-mask-position-ablation}
\setlength{\tabcolsep}{5pt}
\renewcommand{\arraystretch}{1.10}

\begin{tabular}{lcccc}
\toprule
\textbf{Setting}
& \textbf{GSM8K Acc.} 
& \textbf{Token-level $\alpha$} 
& \textbf{Accepted / Block} 
& \textbf{KL(Default $\|$ $\cdot$)} \\
\midrule
With Aligned RoPE
& \textbf{84.0\%}
& \textbf{0.902}
& \textbf{3.61/4}
& \textbf{--} \\
Without Aligned RoPE
& 0.0\%
& 0.718
& 2.87/4
& 7.12 \\
\bottomrule
\end{tabular}

\end{table}
\section{Conclusion}
In this paper, we propose \textbf{SimSD}, a simple but effective plug-and-play strategy that restores token-level temporal structure for supporting dLLMs to use speculative decoding for inference acceleration. By introducing reference tokens from draft-model predictions and controlling their interactions with current-step tokens through a tailored attention mask, our method enables dLLMs to compute valid logits for drafted tokens under temporally consistent contexts. This brings the core verification capability of AR speculative decoding, where multiple drafted tokens are verified in parallel by the target model, to dLLM inference while preserving their parallel generation advantages. Since our method only modifies the attention mask and copies a small number of hidden states, it can be seamlessly integrated into existing dLLM decoding pipelines and combined with acceleration techniques such as KV caching and blockwise decoding.
Extensive experiments on SDAR-family dLLMs across GSM8K, MBPP, TriviaQA, and MMLU demonstrate the effectiveness of our method. Without additional training, it achieves up to \(7.46\times\) higher decoding throughput than the vanilla SDAR baseline with tensor parallelism, while maintaining generation quality and improving average task performance by \(+1.7\%\).

\bibliographystyle{plainnat}
\bibliography{ref}


\end{document}